\documentclass[letterpaper]{article} 
\usepackage[submission]{aaai24}  
\usepackage{times}  
\usepackage{helvet}  
\usepackage{courier}  
\usepackage[hyphens]{url}  
\usepackage{graphicx} 
\usepackage{natbib}  
\usepackage{caption} 
\usepackage{algorithm}
\usepackage{algorithmic}
\usepackage{amsmath}
\usepackage{booktabs}
\usepackage{multirow}
\usepackage{diagbox}

\usepackage{newfloat}
\usepackage{listings}
\DeclareCaptionStyle{ruled}{labelfont=normalfont,labelsep=colon,strut=off} 
\floatstyle{ruled}
\newfloat{listing}{tb}{lst}{}
\floatname{listing}{Listing}
\title{Towards Faster k-Nearest-Neighbor Machine Translation}
\author{
    Xiangyu Shi\textsuperscript{\rm 1},
    Yunlong Liang\textsuperscript{\rm 1},
    Jinan Xu\textsuperscript{\rm 1},
    Yufeng Chen\textsuperscript{\rm 1}
}
\affiliations{
    \textsuperscript{\rm 1}School of Computer and Information Technology, Beijing JiaoTong University\\
    22120416@bjtu.edu.cn, yunlonliang@gmail.com, jaxu@bjtu.edu.cn, chenyf@bjtu.edu.cn

}

\usepackage{bibentry}

\begin{document}

\maketitle

\begin{abstract}
    Recent works have proven the effectiveness of k-nearest-neighbor machine translation(a.k.a kNN-MT) approaches 
    to produce remarkable improvement in cross-domain translations. However, 
    these models suffer from heavy retrieve overhead on the entire 
    datastore when decoding each token. 
    We observe that during the decoding phase,
    about 67\% to 84\% of tokens are unvaried after 
    searching over the corpus datastore, 
    which means most of the tokens cause futile 
    retrievals and introduce unnecessary computational 
    costs by initiating k-nearest-neighbor searches. 
    We consider this phenomenon is explainable in linguistics 
    and propose a simple yet effective multi-layer perceptron (MLP) network to predict 
    whether a token should be translated jointly by the neural machine translation model 
    and probabilities produced by the kNN or just by the neural model. 
    The results show that our method succeeds in reducing redundant retrieval operations and 
    significantly reduces the overhead of $k$NN retrievals by up to 53\%
    at the expense of a slight decline in translation quality. 
    Moreover, our method could work together with all existing kNN-MT systems. 

    \hspace{1em} This work has been accepted for publication in the jornal \textit{Advances in Artificial Intelligence and Machine Learning}\footnote{https://www.oajaiml.com} (ISSN: 2582-9793). The final published version can be found at DOI: https://dx.doi.org/10.54364/AAIML.2024.41111
\end{abstract}

\section{Introduction}

Transformer ~\citep{Transformer} based neural machine translation(NMT) have shown promising performance 
 and evolved various methodologies for further improvements.
 The neural translation models try to learn representations of the semantics on the upstream corpus and 
generalize to downstream tasks by classifying representations to some tokens, 
therefore out-of-domain issues occur. 
The recent concerned $k$ nearest neighbor machine translation($k$NN-MT) ~\citep{khandelwal2021nearest}
proposes a non-parametric method to enhance NMT systems. 
At each decoding phase $k$NN-MT retrieves by current latent representation 
to potential target tokens in a pre-built key-value datastore, 
it jointly considers the probabilities distribution of 
each token given by the neural model and the $k$ nearest reference tokens.
When using an out-of-domain neural translation model, 
the datastore could be created from an 
in-domain corpus, thus $k$NN-MT significantly rises the 
performance on cross-domain machine translations. 
Besides, it could also slightly boost the results of in-domain translations.

\begin{figure}[!t] 
    \centering 
    \includegraphics[width=0.9\columnwidth]{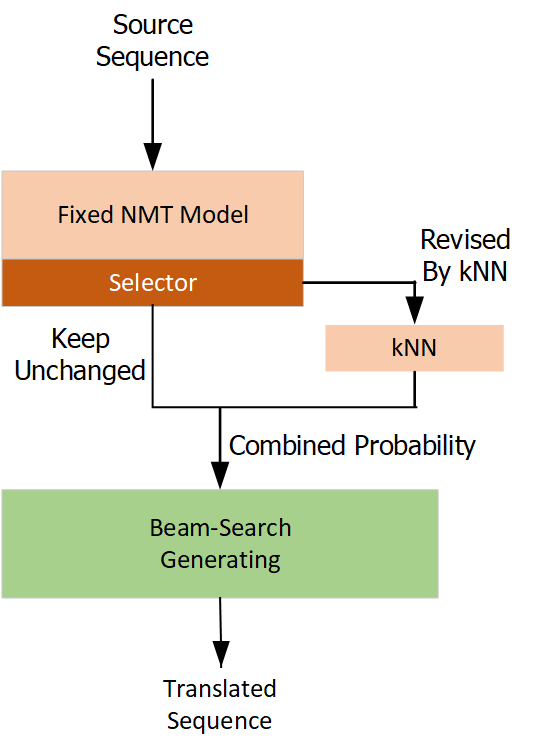} 
    \caption{Overview of our method. The selector predicts all the tokens to decide whether to revise the probability distribution. $k$NN retrievals only occur on parted tokens instead of all tokens, thus we speed up the $k$NN-MT systems.} \label{fig:inference}
\end{figure}

However, massive retrievals from entire datastore 
by high-dimension representations bring
 non-negligible computational costs to $k$NN-MT systems. 
 According to the primitive paper of $k$NN-MT~\citep{khandelwal2021nearest}, 
 $k$NN introduces two-orders slower to original NMT models.
 This overhead limits the applications of $k$NN-MT systems 
  to some practical scenarios, such as instant translation tasks.

\begin{table}[t]
\centering
\begin{tabular}{l c c c c}
    \toprule
    & IT & Koran & Law & Medical \\
    \midrule 
    Redundant & 0.80 & 0.67 & 0.84 & 0.81 \\ 
    \bottomrule
\end{tabular}
\caption{Redundant ratio tested on multi-domain dataset
~\citep{Koehn2017six,aharoni-goldberg-2020-unsupervised}. The results are measured on test set, and beam searching has been considered in. }
\label{tb:redundant_ratio}
\end{table}

Previous works for accelerating $k$NN-MT systems suggest
 shrinking the size of the datastore~\citep{meng-etal-2022-fast,wang-etal-2022-efficient,martins-etal-2022-efficient},
 reducing the dimension of representation for retrieval~\citep{wang-etal-2022-efficient,martins-etal-2022-efficient}.
 We observe that about 67\%-84\% predicted tokens 
 are kept unchanged on its datasets after being revised by the vanilla $k$NN-MT system, which means retrievals for these tokens 
 bring redundant computational overhead, as shown in table ~\ref{tb:redundant_ratio}. 
 Towards faster $k$NN-MT,
 we proposed a simple yet effective MLP to predict the necessity to retrieve each token. 
 The translation procedure is shown as figure ~\ref{fig:inference}. We only add an MLP (so-called selector) after an off-the-shelf NMT model. 

 We propose our contributions are:

 \begin{itemize}
    \item We observe that most of the tokens do not require retrievals. This opens up a new idea for $k$NN-MT.
    \item We provide a simple baseline model for this idea.
    \item Our method and potential similarity methods could work together with any existing $k$NN-MT system for further accelerating.
 \end{itemize}

\section{Background}

This section briefly introduces the background of $k$NN-MT, including its previous works, methodologies, and diverse $k$NN-MT variants. 

\subsection{Neural Machine Translation}

Currently, neural machine translation is implemented under sequence to sequence~\citep{NIPS2014_seq2seq} 
frameworks with attench mechanism~\citep{mnih2014recurrent, Bahdanau2015Neural}.
 Given a source sequence $\textbf{x} = \lbrace x_1, \cdots, x_n \rbrace$, the NMT model $\mathcal{M}$ translates $\textbf{x}$ to 
the target sequence $\textbf{y} = \lbrace y_1, \cdots, y_m \rbrace$ in another language. 
At each step of decoding, 
the NMT model produces the probability distribution over the vocabulary $p_{MT}(\hat{y_i}|\textbf{x},\hat{\textbf{y}}_{1:i-1})$. This probability will be used 
in beam search for text generation. 
NMT models trained on particular domains perform deterioration when 
translating out-domain sentences\citep{Koehn2017six,chu-wang-2018-survey}.

\subsection{Domain Adaptation for NMT}

The most concerned approach to adapt a general domain NMT model to a partial domain is to continue 
training the neural model on the in-domain corpus. However, this finetuning method 
demands large computational resources and suffers from the notorious catastrophic forgetting issue~\citep{MCCLOSKEY1989109, santoro2016oneshot}.
Moreover, in real applications scenarios, the domains of translating
sentences are rarely known ahead of the time. 
So multi-domain neural machine translation in one architecture is proposed~\citep{farajian-etal-2017-multi, pham-etal-2021-revisiting,lin-etal-2021-learning}.

\subsection{Retrieval-Argmented Text Generation}  

Retrieval-augmented methods are concerned in text generation tasks recently.
RetNRef ~\citep{weston-etal-2018-retrieve}
concatenate the representations of a generative LSTM with attention model and the embeddings of retrieved over dialogue history, then use it to generate as usual.
Knowledge-intensive question answering is augmented by retrieving similar documents from Wikipedia
as contexts~\citep{NEURIPS2020_6b493230}. BERT-$k$NN ~\citep{kassner-schutze-2020-bert} 
adds a kNN searching over a large datastore after original BERT model~\citep{devlin-etal-2019-bert} 
and excels the BERT model on question answering by large margin. 
BERT-$k$NN model gives more precise answers than the baseline model and can learn new knowledge from the datastore without training.
As for machine translation, 
Gu et al. ~\citep{gu2018search} propose to retrieve several 
parallel sentence pairs based on edit distance with source sequence to perform the translation.

The common pattern is to retrieve similar texts or embeddings in a datastore and use retrieving results to enhance generation.
Training a tuned neural translation model requires enormous computing power and a parallel corpus.
Non-parametric or few-parametric retrieval-based approaches offer opportunities to learn new knowledge and adapt to new domains at a low cost.

\subsection{$k$NN-MT}

The basis of $k$NN-MT is datastore creation and retrieving for probability distributions.

\subsubsection{Datastore} $k$NN-MT first creates a datastore on corpus $\mathcal{C}$, for each parallel sequence pair $(\textbf{x}, \textbf{y})$,
 it adds several key-value pair 
 $\lbrace (f(\textbf{x},\textbf{y}_{1:i-1}), y_i) | i=1,\cdots,m \rbrace$ into the datastore $\mathcal{D}$,
  where $f(\textbf{x},\textbf{y}_{1:i-1})$ is the intermediate representation given by the decoder of $\mathcal{M}$.
  Therefore, the datastore could be created on any parallel corpus without any training step.

\subsubsection{Estimation of Probability Distribution} At decoding time given input $(\textbf{x},\hat{\textbf{y}}_{i-1})$, 
it makes a query $q = f(\textbf{x},\hat{\textbf{y}}_{i-1})$ and retrieves $k$ nearest 
neighbor $k_1,\cdot, k_K$ with their corresponding target tokens. Thus, $p_{kNN}$ is estimated by $L_2$ distances $d$ and a temperature $T$, 
normalizing retrieved set into a probability distribution over vocabulary by softmax:

\begin{equation} \label{eq:pkNN}
 p_{KNN}(\hat{y_i}|\textbf{x},\hat{\textbf{y}}_{i-1}) = \mathrm{sofmtax}(\frac{-d(q, k_j)}{T}), j=1,\cdots, K 
\end{equation}

Temperature $T$ greater than one flattens the distribution and prevents overfitting to the only nearest retrieval~\citep{khandelwal2021nearest}. 

The probability distribution given by the NMT model and $k$NN is finally interpolated with a hyper-parameter $\lambda$:

\begin{equation} \label{eq:revised_prob}
\begin{aligned} 
p_{\text{\textit{combined}}}(\hat{y_i}|\textbf{x},\hat{\textbf{y}}_{i-1}) &= \lambda p_{KNN}(\hat{y_i}|\textbf{x},\hat{\textbf{y}}_{i-1})  \\
        & + (1-\lambda)p_{MT}(\hat{y_i}|\textbf{x},\hat{\textbf{y}}_{1:i-1})
\end{aligned} 
\end{equation}

\subsection{Adaptive-kNN-MT}

Adaptive variant~\citep{zheng-etal-2021-adaptive} adds a light-weight feed-forward network named \textit{Meta-k} network 
to dynamically assign weights for probability distributions given by the neural model and $k$ nearest reference sample, the final prediction is obtained by 

\begin{equation}
\begin{aligned}
    p_{\text{\textit{combined}}}(\hat{y_i}|\textbf{x},&\hat{\textbf{y}}_{i-1})\\ 
    &= p_{\text{Meta}}(f(\textbf{x},\textbf{y}_{1:i-1})) \cdot p_{MT} (\hat{y_i}|\textbf{x},\hat{\textbf{y}}_{1:i-1}) \\
    &+ \sum_{j=1}^{K} p_{\text{Meta}}(k_j) \cdot p_{k_iNN}(\hat{y_i}|\textbf{x},\hat{\textbf{y}}_{1:i-1}) \notag
\end{aligned}
\end{equation}

Here $p_{\text{Meta}}$ is calculated by a simple feed-forward network with distances and distribution of target tokens of its $k$ nearest neighbors,
 $p_{k_iNN}$ is also given by Eq. ~\ref{eq:pkNN}.
The Adaptive approach introduces almost no inference latency and achieves 1.44 $\sim$ 2.97 BLEU score improvements than the vanilla $k$NN-MT.

\subsection{PCK-kNN-MT}

PCK variant~\citep{wang-etal-2022-efficient} extends the Adaptive method and 
trains a compact network to reduce the dimension of key vectors. 
It also makes margins between keys with different target tokens more discriminable. 
Then it prunes the size of the datastore by deleting redundant key-value pairs. 
It is a typical and effective accelerating method of reducing the query dimension and shrinking the datastore.
Compared with our method, which accelerates $k$NN-MT by eliminating unnecessary retrievals and is simple also highly interpretable. 

We will show that our method could work with the Adaptive and the PCK $k$NN-MT.

\subsection{Some Other $k$NN-MT Variants}

\begin{itemize}
    \item \textbf{SK-MT}~\citep{dai2023simple}, a distance-ware adapter is introduced to adaptively incorporate retrieval results.
    \item \textbf{Revised Key KNN-MT}~\citep{cao-etal-2023-bridging}, a simple feed-forward netowrk is trained to add a revising vector to original query. 
                It improves the retrieving accuarcy and enhances the adaptation.  
    \item \textbf{Robust KNN-MT}, introducing NMT condfidence to alleviate deterioration casued by noisy key-value pairs.
\end{itemize}

\subsection{kNN-BOX}

kNN-BOX~\citep{zhu2023knnbox} assembles various implementations of $k$NN-MT systems, together with the baseline NMT model. It provides a unified 
framework for modeling and evaluations. It uses FAISS~\citep{johnson2019billion} for efficient high-dimensional vector retrievals implementations. 
 Our implementation and experiments are based on this framework.
 It is a remarkable fact that kNN-BOX further optimizes the codes and significantly speeds up retrievals, 
 and we do not observe two-orders slower in speed as the original paper reports.

\section{Methodolgy}

\begin{figure}[h] 
    \centering 
    \includegraphics[width=0.98\columnwidth]{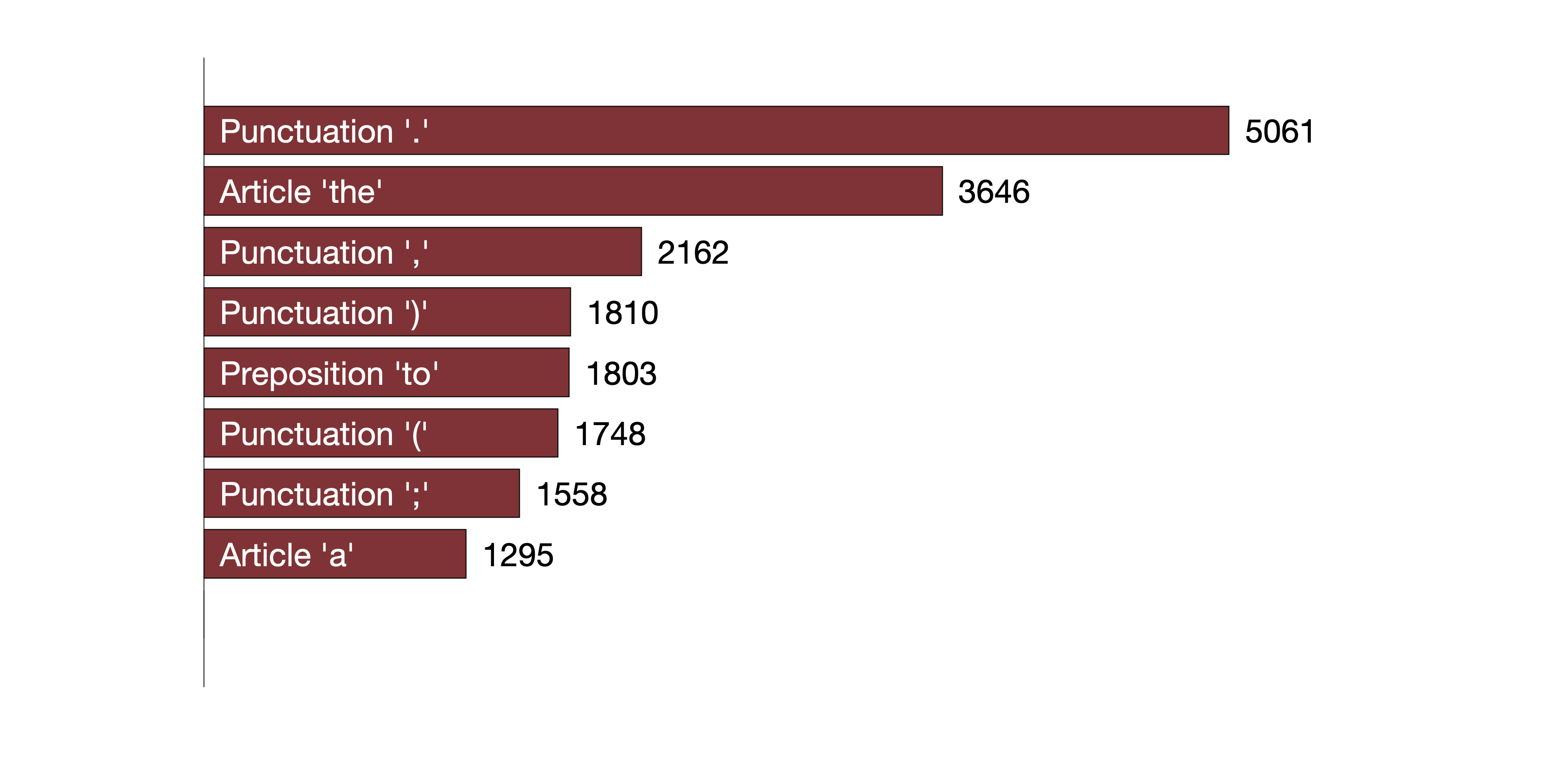} 
    \caption{Top-8 tokens which cause futile retrievals on multi-domain IT dataset. 
    These tokens are always unchanged after $k$NN retrievals. } \label{fig:top5}
\end{figure}

We consider that the reason why most tokens do not require $k$NN retrieval in cross-domain 
machine translation is that the frequent words always keep unvaried in diverse domains, for example, in English, they may be punctuations and prepositions. Meanwhile, the out-of-domain issue often 
occurs in nouns and verbs.
We count these types of words, and the ratios keep unchanged after $k$NN retrievals, for example, as shown in figure ~\ref{fig:top5}
Thus, the idea is highly explainable.

\subsubsection{Our Mehtod} We train a simple 3-layer MLP network called selector  
to predict whether each token to retrieve or not.
We want the selector directly distinguish the latent representations of in-domain and out-domain semantics.
 It gives the probabilities for the two options by 

\begin{equation}
\begin{aligned}
    p(\mathcal{A}_i|f&(\textbf{x}, \textbf{y}_{1:i-1})) \\ = 
    & \mathrm{softmax}(W_2^T [\mathrm{ReLU}(W_1^T \cdot f(\textbf{x},\textbf{y}_{1:i-1}))])
\end{aligned}
\end{equation} 
where $W_1 \in \mathcal{R}^{d \times d'}$ and $W_2 \in \mathcal{R}^{d' \times 2}$. 
$d'$ is the inner hidden dimension of the MLP. 
$\mathcal{A}_i \in \lbrace 0, 1 \rbrace$ is the decision of whether to retrieve for $y_i$, 
0 is to retrieve and 1 is not.

\begin{figure}[h] 
    \centering 
    \includegraphics[width=0.86 \columnwidth]{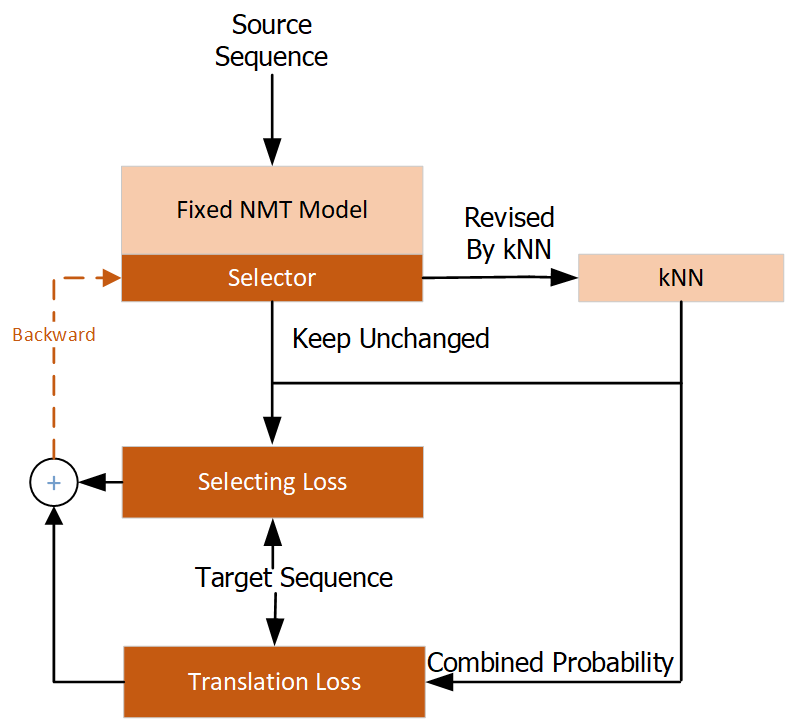} 
    \caption{The training procedure of our method.} \label{fig:train}
\end{figure}

\subsubsection{Training Procedures} 
We make labels $l$ for the selector by observing the 
predictions of the neural
translation model $\mathcal{M}$ and the target sequences $\textbf{y}$. 
If $\hat{y}_i = \mathcal{M}(f(\textbf{x},\textbf{y}_{1:i-1}))$ is equal to $y_i$, 
we mark the label of $f(\textbf{x},\textbf{y}_{1:i-1})$ to be 1, otherwise 0. 
Weighted cross-entropy loss is used as the criterion to train the network, the loss is measured by 

\begin{equation}
\begin{aligned} \label{celoss}
\mathcal{L}_1 & = \sum_{i} [ -\frac{N}{B} [l_i=0] \log p(\mathcal{A}_i=0|f(\textbf{x}, \textbf{y}_{1:i-1}))   \\
              & -(1-\frac{N}{B}) [l_i=1] \log p(\mathcal{A}_i=1|f(\textbf{x}, \textbf{y}_{1:i-1})) ]
\end{aligned}
\end{equation}

In Eq.~\ref{celoss} $B$ is the number of the tokens in the training batch, and $N$ is the number of negative samples where $\mathcal{A}_i = 1$

The selector predicts each token and impacts the decoding results, 
so we also want to train the network by translation loss. For each sample, 
the selector selects those tokens requiring retrievals then revises their probability distributions by Eq. ~\ref{eq:revised_prob}, 
the rest tokens keep their probability distributions unvaried. The final probability distribution over vocabulary $V$ for $\hat{y}_i$ is:

\begin{equation}
    p(\hat{y}_i |\textbf{x},\hat{\textbf{y}}_{1:i-1} ) = 
    \left\{\begin{matrix}
        p_{MT}(\hat{y}_i |\textbf{x},\hat{\textbf{y}}_{1:i-1})&  \mathcal{A}_i = 0 \\
        p_{\text{\textit{combined}}}(\hat{y}_i |\textbf{x},\hat{\textbf{y}}_{1:i-1}) & \mathcal{A}_i = 1 
    \end{matrix}\right. 
\end{equation}
    
Thus we could train the parameters by translation loss: 

\begin{equation}\label{eq:mtloss}
    \mathcal{L}_2 = \sum_{i} [ -\sum_{\hat{y}_i \in V} [y_i == \hat{y}_i] \log p(\hat{y}_i | \textbf{x},\hat{\textbf{y}}_{1:i-1} ) ]
\end{equation}

And the final loss:

\begin{equation}
    \mathcal{L} = \frac{\mathcal{L}_1}{N} + \frac{\mathcal{L}_2}{N}
\end{equation}

Where N is the total number of tokens in the training batch, the entire training procedure is shown in figure ~\ref{fig:train}.

Another problem is to determine 

\begin{equation}
\mathcal{A}_i = \mathrm{argmax}_{\mathcal{A}} p(\mathcal{A}|f(\textbf{x},\textbf{y}_{1:i-1})), \mathcal{A} \in \lbrace 0,1 \rbrace 
\end{equation}

Undifferentiable argmax operation is used to address gradient intercept issues, 
we use the gumbel-softmax~\citep{jang2016gumbelsoftmax} trick to approximate a gradient for the argmax operation as follows:

\small
\begin{equation}
\begin{aligned}
& G \approx \\ & \nabla_W \frac{ \exp((\log p(\mathcal{A}|f(\textbf{x},\textbf{y}_{1:i-1}); W) ) + g_m(\mathcal{A}'))/\tau ) }{\sum\limits_{\mathcal{A}' \in \lbrace 0,1 \rbrace} \exp((\log p(\mathcal{A}'|f(\textbf{x},\textbf{y}_{1:i-1}); W) ) + g_m(\mathcal{A}'))/\tau)  } \notag
\end{aligned}
\end{equation}
\normalsize

Where $W$ is the inner trainable parameters of the selector, 
and $g_m = -\log(-\log(u))$ with $u \sim U(0,1)$, $\tau$ is the temperature hyper-parameter.

\subsubsection{Metrics} 

To evaluate our method, we use the following metrics. 
The time is measured three times and averaged. 

\begin{itemize}
    \item \textbf{Sacre-BLEU} ~\citep{post-2018-call} is the recommended by WMT to be the evaluating metrics.
    \item \textbf{Inference time}: The total time consumed for the neural model, selector(if exists) and $k$NN to translate all testing sequences. 
    \item \textbf{KNN overhead time}: A time composed of selector prediction, $k$NN retrievals and probability distribution revision. It is also a part contained in the total inference time. 
        The reason why we do not only use inference time or tokens per second metrics but also measure this indicator is discussed later.
    \item \textbf{Tokens per second} The average tokens translated per second.
    \item \textbf{Precision} and \textbf{Recall}: We dictate that samples requiring retrievals are positive and others are negative. Therefore higher precision index indicates less futile retrievals and is closer to the speed of the pure NMT model. A high recall index means approaching full retrievals on $k$NN and better translation quality.
    \item \textbf{Ratrieving ratio}: Besides, we introduce a new metric retrieving ratio for the selector, it is the ratio of tokens predicted not to retrieve to the total number of tokens.
\end{itemize}




\subsubsection{Integration}

Our selector could be integrated into any other 
$k$NN-MT systems by training the selector and 
save its weights on vanilla $k$NN-MT then loading it in any $k$NN-MT variants. 
The procedure could be described as pseudo shown in algorithm ~\ref{alg:algorithm}.

\begin{algorithm}[tb]
    \caption{Integrating the selector into $k$NN-MT systems}
    \label{alg:algorithm}
    \textbf{Input}: Source sequnece $\textbf{x}$ and partial translated $\hat{\textbf{y}}_{1:i-1}$  \\
    \textbf{Parameter}: NMT Model $\mathcal{M}$, Selector $\mathcal{S}$, Datastore $\mathcal{D}$, $K$ \\
    \textbf{Output}: $p(\hat{y}_i |\textbf{x},\hat{\textbf{y}}_{1:i-1} )$
    \begin{algorithmic}[1] 
        \STATE Let $z = f_\mathcal{M}(\textbf{x}, \hat{\textbf{y}}_{1:i-1})$
        \STATE Let $p_{MT}(\hat{y_i}|\textbf{x},\hat{\textbf{y}}_{1:i-1}) = \mathcal{M}(z)$ 
        \IF {$\mathcal{S}(z) == 0$}
            \STATE Retrieve $K$ nearest neighbors from $\mathcal{D}$ \\ 
            \STATE Estimated $p_{kNN}(\hat{y_i}|\textbf{x},\hat{\textbf{y}}_{1:i-1})$ by Eq. ~\ref{eq:pkNN}
            \STATE Caculate $p_{combined}(\hat{y_i}|\textbf{x},\hat{\textbf{y}}_{1:i-1})$ by Eq. ~\ref{eq:revised_prob}
            \STATE \textbf{return} return $p_{combined}(\hat{y_i}|\textbf{x},\hat{\textbf{y}}_{1:i-1})$
        \ELSE 
            \STATE \textbf{return} $p_{MT}(\hat{y_i}|\textbf{x},\hat{\textbf{y}}_{1:i-1})$
        \ENDIF
    \end{algorithmic}
\end{algorithm}

\begin{table}[h]
    \centering
    \begin{tabular}{l c c c c}
        \toprule
        & IT & Koran & Law & Medical \\
        \midrule 
        $\lambda$ & 0.7 & 0.8 & 0.8 & 0.8 \\ 
        T & 10 & 100 & 10 & 10 \\
        \bottomrule
    \end{tabular}
    \caption{These values come from the original $k$NN-MT paper~\citep{khandelwal2021nearest} figure 5.  }
    \label{tb:hpprarm_1}
\end{table}

\begin{table*}[t]
    \centering
    \begin{tabular}{lcccccccc|cc}
        \toprule
        \multirow{2}*{Systems}  & \multicolumn{2}{c}{IT} & \multicolumn{2}{c}{Koran} & \multicolumn{2}{c}{Law} & \multicolumn{2}{c}{Medical} & \multicolumn{2}{c}{Averge}  \\
        \cmidrule{2-11}
          &                        Total   & KNN  & Total & KNN & Total  & KNN  & Total & KNN & Total & KNN \\
        \midrule
    

        Vanilla KNN-MT    &   18.83 & 4.04  &   28.90 & 5.02 & 64.53 &  22.22  &  39.33 & 9.43 &  37.90 & 10.18 \\
        \hline

        \multirow{2}*{Vanilla + Selector}  &   \multirow{2}*{16.77}  & 2.41  &  \multirow{2}*{27.77} & 3.74 & \multirow{2}*{52.6} & 10.38  &  \multirow{2}*{34.63} & 5.97 &  \multirow{2}*{32.94} & 5.63 \\

        & & (-40.3\%) & & (-34.2\%) & & (-53.3\%) & & (-36.7\%) & & (-44.7\%)\\
        \bottomrule
    \end{tabular}
    \caption{ Time comparison in seconds. For each dataset we measured the inference time(Total) and $k$NN overhead time(KNN) metrics.
    To measure the exact time, the global CUDA stream is synchronized before the boundary of the relevant code snippets, slightly reducing operational efficiency.
    KNN overhead time comprises $k$NN retrieving, selector predicting, and probability distribution revising. 
    The server equips with an Intel Core i9-12900K CPU and an Nvidia Quadro RTX8000 GPU. } \label{tb:results_2}
\end{table*}

\begin{table*}[t]
    \centering
    \begin{tabular}{lcccccccc|cc}
        \toprule
        \multirow{2}*{Systems}  & \multicolumn{2}{c}{IT} & \multicolumn{2}{c}{Koran} & \multicolumn{2}{c}{Law} & \multicolumn{2}{c}{Medical} & \multicolumn{2}{c}{Averge}  \\
        \cmidrule{2-11}
          &                         BLEU  & \small{Tokens/s}  & BLEU & \small{Tokens/s} & BLEU  & \small{Tokens/s}  & BLEU & \small{Tokens/s} & BLEU & \small{Tokens/s} \\
        \midrule
        
        Pure NMT Model      &     38.35  & 2171.8  &   16.26 & 2199.0 & 45.48 &  1922.6  &  40.06 & 1919.6 &  35.04 & 2053.3 \\     
        
        \hline 
        \multicolumn{11}{c}{Vanilla, $K$ = 8} \\
        \hline 

        Vanilla KNN-MT      &     45.52  & 1638.3  &   20.54 & 1720.8 & 61.08 &  1243.0  &  53.51 & 1347.2 &  45.16 & 1487.3 \\
       
        Vanilla + Selector  &     43.90  & 1843.1  &   18.55 & 1820.2 & 57.18 &  1509.1  &  48.70 & 1540.4 &  42.08 & 1678.2 \\
        \hline
        \multicolumn{11}{c}{Adaptive, maximum $K$ = 4} \\
        \hline
        Adaptive KNN-MT     &     47.78  & 1676.0  &   20.23 & 1749.3 & 63.00 &  1230.1  &  56.31 & 1371.4 &  46.83 & 1506.7 \\
        Adaptive + Selector &     46.12  & 1857.2  &   18.64 & 1847.0 & 58.71 &  1489.8  &  50.26 & 1619.5 &  43.43 & 1703.4 \\
        \hline 
        \multicolumn{11}{c}{PCK, maximum $K$ = 4} \\
        \hline
        PCK KNN-MT          &     47.61  & 1774.5  &   19.74 & 1788.0 & 62.95 &  1376.0  &  56.62 & 1510.1 &  46.73 & 1612.2 \\
        PCK + Selector      &     45.77  & 1891.8  &   18.67 & 1869.5 & 57.96 &  1552.2  &  50.38 & 1583.0 &  43.20 & 1724.1 \\
        \bottomrule
    \end{tabular}
    \caption{ Domain adaption performance on the test dataset. For the PCK method, we do not perform a datastore prune to keep the same datastore scale. 
    So our selector remarkably improves translation speed for all these $k$NN-MT variants.
    The larger the size of the datastore and the number of neighbors $K$, there is the more pronounced the advantage of reducing redundant retrievals. } \label{tb:results1}
\end{table*}

\begin{table}[h]
    \centering
    \begin{tabular}{l c c c}
        \toprule
          & Precision & Recall & Retrieving Ratio \\
        \midrule 
        IT & 0.60 & 0.81 & 0.53 \\ 
        Koran & 0.70 & 0.78 & 0.58 \\
        Law & 0.53 & 0.81 & 0.48 \\
        Medical & 0.64 & 0.80 & 0.49 \\
        \bottomrule
    \end{tabular}
    \caption{The metrics of the selector.}
    \label{tb:selector_performance}
\end{table}

\begin{table}[h]
    \centering
    \begin{tabular}{l c c c}
        \toprule
          & Precision & Recall & Retrieving Ratio \\
        \midrule 
        IT       & 0.68(+0.08) & 0.60(-0.21) & 0.35(-0.18) \\ 
        Koran    & 0.76(+0.06) & 0.62(-0.16) & 0.42(-0.16) \\
        Law      & 0.58(+0.05) & 0.73(-0.08) & 0.39(-0.09) \\
        Medical  & 0.67(+0.03) & 0.77(-0.04) & 0.44(-0.05) \\
        \bottomrule
    \end{tabular}
    \caption{The metrics of the selector trained without translation loss. (+/-*) compares with counterparts in table ~\ref{tb:selector_performance}.}
    \label{tb:selector_wo_tl}
\end{table}

\begin{table}[h]
    \centering
    \begin{tabular}{c c c c}
        \toprule
          IT & Koran & Law & Medical \\
        \midrule 
          41.53\small{(-2.37)} & 17.91\small{(-0.64)} & 54.90\small{(-2.28)} & 47.59\small{(-1.11)}  \\
        \bottomrule
    \end{tabular}
    \caption{ BLEU scores of vanilla $k$NN-MT involving a selector trained without translation loss. (-*) compares with counterparts in table ~\ref{tb:results1} on the row 'Vanilla + Selector'.}
    \label{tb:trans_quality_wo_tl}
\end{table}

\section{Experiments}

\subsection{Hyper-parameters}

Adopting the WMT19 En-De model ~\citep{Natha2020WMT19de-en} and freezing its parameters during training, We employ the crucial hyper-parameters interpolating index $\lambda$ and temperature $T$ the same values in the vanilla $k$NN-MT as table ~\ref{tb:hpprarm_1}

As for $k$NN-MT systems with adaptive \textit{Meta-k} network, we set the maximum number of neighbors to 4, 
and as for others, we form to retrieve fixed 8 nearest neighbors. 

We create datastore on the training corpus and train the selector on the valid dataset 
of each domain as $k$NN-BOX framework used to do. An Adam optimizer ~\citep{kingma2014adam} is used for training 
100 epochs with an initial learning rate to be 1e-4. 
Temperature $\tau$ for Gumbel-softmax is fixed at 0.1, 
more details are reported in the appendix.

\subsection{Results}

\subsubsection{The selector succeeds in discriminating samples that need retrievals.} 
Table ~\ref{tb:selector_performance} shows the classifying quality of the selector. 
These metrics indicate that the selector selects about 50\% tokens to retrieve, and about 80\% of them genuinely require retrievals. By contrast, if each sample is randomly 
predicted, the Recall metric should be only about 50\%. According to table ~\ref{tb:redundant_ratio},
only 16\%-33\% tokens entail retrievals, but the selector performs much higher Precision metrics, 
The results have proven the selector to be effective. Our approach is independent of both retrieving strategy and the datastore. Experiments only test the $k$NN overhead time reduction after adding the selector to the Vanilla $k$NN-MT, but in theory, 
there should also be speedups close to table ~\ref{tb:results_2} for any other $k$NN-MT systems.

\subsubsection{The selector significantly speeds up existing $k$NN-MT system.} 

Table ~\ref{tb:results_2} indicates the selector optmizes 36.7\% to 53.3\% $k$NN overhead time.
Table ~\ref{tb:results1} represents the translation qualities and inference speed.
With a selector, all other $k$NN-MT systems gain accelerated.

\subsubsection{Ablation for translation loss.}

We also train selectors without translation losses described by Eq. ~\ref{eq:mtloss}, 
Note that Gumbel-softmax trick is turned off when translation losses are undesired.
the results are present in table ~\ref{tb:selector_wo_tl} and table ~\ref{tb:trans_quality_wo_tl}.
We observe considerably lower translation quality on all the datasets, 
however, remarkable variations of selector metrics occur on the IT and Koran dataset 
 while slight changes in selector metrics occur on Law and Medical dataset. 
It is unclear how these metrics quantitatively affect translation quality,
but we can empirically conclude that translation losses do not, at the very least, 
lead to model deterioration.

\section{Discussion}

\subsubsection{Different implementations of baselines from previous works}

It can be seen from table ~\ref{tb:results_2} that $k$NN retrievals only take up a small 
part of the time in inference time, one reason is that we do not use too large datastore due to hardware limitations,
the other reason is that we adopt the KNN-BOX framework, which remarkably optimizes the retrieving performances as the baseline code implementations,
rather than comparing the results with the code of the original $k$NN-MT paper. 
Two-orders slower is reported in that paper, 
which means $k$NN retrievals take up almost all the inference time,
therefore, based on the code of the original $k$NN-MT, the improvements of metrics such as inference time or tokens per second 
can be completely considered as an improvement from their methods,
but in our case, we could not do so. Therefore we measure and compare $k$NN overhead times.

\subsubsection{Comments on our method}
Although the selector performs pretty, there is still room for improvement. 
The idealized result is to distinguish all tokens that truly need retrievals and make no futile retrieval.
This motivation and mechanism are clear and explainable, rule-based and statistics-based techniques may help.

Reducing undesired retrievals is a promising research direction for fast $k$NN-MT.

Due to unbalanced positive and negative samples, complex and dense latent representations, 
it is hard for a simple MLP network to learn strong discrimination, 
while more complex models take more time. 

These efforts have been applied to further improve the selector, but they do not make a difference.

\begin{itemize}
    \item Use the probability distribution $p_{MT}(\hat{y_i}|\textbf{x},\hat{\textbf{y}}_{1:i-1})$ given by $\mathcal{M}$ 
        as the input of the selector, which greatly extends the number of parameters of the selector(about 40 times) and does not work.
    \item Add more one hidden layer in the selector. 
    \item Train the selector on the train set trying to enhance generalization, but it fails to improve metrics. 
    \item Use DiceLoss~\citep{li2019dice} and FocalLoss~\citep{lin2017focal} instead of weighted cross-entropy loss. 
    They cause more fearful overfitting issues, the selector predicts nearly all the tokens to retrieve.
\end{itemize}

We do not rule out that the last two points are caused by inappropriate hyper-parameters or lack of training ticks.

\section{Conclusion}

This paper opens a new idea for faster $k$NN-MT. 
We propose a simple yet effective selector to 
reduce redundant retrievals. Experiment results on four 
benchmark datasets show that the selector remarkably speeds up 
other $k$NN-MT systems and keeps acceptable translation qualities. 
The limitation is that it could not outperform the depending $k$NN-MT system in translation quality.
Future work could further investigate the characteristics
of hidden representations from neural models of out-domain tokens to identify out-domain features better.

\bibliography{aaai24}


\end{document}